\documentclass[10pt,twocolumn,letterpaper]{article}

\usepackage{mysiggraph}

\cvprfinalcopy % *** Uncomment this line for the final submission

\def\cvprPaperID{2} % *** Enter the CVPR Paper ID here

\ifcvprfinal\fi
\begin{document}

%%%%%%%%% TITLE
\title{Deep Octree-based CNNs with Output-Guided Skip Connections \\ for 3D Shape and Scene Completion}

\author{Peng-Shuai Wang\\
    Microsoft Research Asia\\
    {\tt\small penwan@microsoft.com}
    \and
    Yang Liu\\
    Microsoft Research Asia\\
    {\tt\small yangliu@microsoft.com}
    \and
    Xin Tong\\
    Microsoft Research Asia\\
    {\tt\small xtong@microsoft.com}
  }

\maketitle
%\thispagestyle{empty}
%
%\twocolumn[{%
%\maketitle
%\vspace{-0.75cm}
%\renewcommand\twocolumn[1][]{#1}%
%\begin{center}
%   \centering
%   \includegraphics[width=1\textwidth]{teaser_v4}
%   \captionof{figure}{3D shape and scene completion by our deep network.
%   The input point clouds are shown in the odd columns, and the output results are
%   shown in the even columns.
%   The meshes in the first row are Poisson reconstructed on the
%   complete point cloud generated by our network.
%   The scenes in the second row are in a voxelized representation and the
%   colors represent different semantic labels.}
%   \label{fig:teaser}
%\end{center}%
%}]

%%%%%%%%% ABSTRACT
\begin{abstract}
Acquiring complete and clean 3D shape and scene data is challenging due to geometric occlusion and insufficient views during 3D capturing.
We present a simple yet effective deep learning approach for completing the input noisy and incomplete shapes or scenes.
Our network is built upon the octree-based CNNs (O-CNN)  with U-Net like structures, which enjoys high computational and memory efficiency and supports to construct a very deep network structure for 3D CNNs. A novel output-guided skip-connection is introduced to the network structure for better preserving the input geometry and learning geometry prior from data effectively.  We show that with these simple adaptions --- output-guided skip-connection and deeper O-CNN (up to 70 layers),  our network achieves state-of-the-art results in 3D shape completion and semantic scene computation.
\vspace{-4mm}
\end{abstract}

%%%%%%%%%%%%%%%%%%%%%%%%%%%%%%%%%%%%%%%%%%%%%%%%%%%%%%%%%%%%%%%%%%%
\section{Introduction}
\label{sec:intro}

Despite the rapid development in 3D capturing techniques, it is still challenging to acquire accurate and complete 3D shapes and scenes due to the interference by shape geometry, surface material, lighting conditions as well as the noise introduced in the capturing process. Therefore, recovering a complete and accurate 3D geometry from partial and noisy
3D inputs become an essential task in 3D acquisition.

3D completion is inherently an ill-conditioned problem. Many methods
have been proposed to tackle this challenging problem based on different priors.
Optimization-based methods~\cite{Kazhdan2006,Calakli2011,Andrea2011} exploit the
local geometry properties, \eg smoothness of the local surface or volume, for 3D shape completion.
Although these methods are able to fill small holes well, they cannot
recover large missing regions. Matching-based methods
~\cite{Shen2012,Shao2012} reconstruct 3D shapes with the help of surfaces or parts
found in a 3D shape database that best match the input partial shape.
However, these methods are sensitive to noise and could fail if no similar shapes exist in the database. Recently, learning-based methods
have been proposed for 3D shape completion~\cite{Han2017,Dai2017,Wang2018a,Wu2018b,Yang2018}
and scene completion~\cite{Song2017,Dai2018,Zhang2019,Liu2018a,Wang2019}.
Inspired by the learning techniques on the image domain, many dedicated network structures, and various loss functions have been designed for learning a general yet compact latent space from 3D data to infer complete 3D geometry.
However, their direct and na\"{\i}ve extension from 2D images to 3D voxels introduces high-memory cost and inefficient computation issues. Limited by this inefficient 3D representation, many existing 3D learning methods are still in a shallow network architecture and have not benefited from the power of \emph{deep} layers, which proved extremely useful for 2D vision and NLP learning~\cite{Goodfellow2016}.

In this paper, we present a deep learning approach for 3D shape and scene completion.
Taking a noisy and incomplete point cloud of a 3D shape or scene as input, our method
represents the input with an efficient octree structure and predicts the complete output via deep octree-based CNNs with novel \emph{output-guided skip connections}. Our deep octree-based CNNs are based on the O-CNN framework~\cite{Wang2017,Wang2018a} which is highly efficient both in memory and computation cost and makes deep layers possible. Our network design for 3D completion follows the U-Net structure~\cite{Ronneberger2015}, which consists
of two deep residual networks~\cite{He2016} for encoding and decoding.
The encoding network is defined on the input octree and transforms the input into
a compact latent code, while the decoding network takes the latent code as input
to infer the octree and detailed point cloud of the complete shape or scene.
As the input and output octrees are different due to this complete task,
not all features defined at one octree level of the decoding
network can find the corresponding features at the same octree level of the encoding
network. We propose output-guided skip connections that add skip connections between the generated octree node and its corresponding and existing octree node in the input octree only.
This output-guided scheme well preserves the geometric information in the input and is robust to the input noise.

We evaluate the effectiveness of our approach in typical benchmarks of 3D shape completion and
semantic scene completion tasks. Experiments show that our simple network design --- efficient 3D representations based on octree, deep layers, and output-guided skip connections, outperforms the existing approaches and achieves state-of-the-art results.

%We believe our work will help to simplify the 3D network design and avoid ad-hoc network
%architectures for many 3D learning tasks.

\section{Related Work}
\label{sec:related}

%%%%%%%%%%%%%%%%%%%%%%%%%%%%%%%%%%%%%%%%%%%%%%%%%%%%
\paragraph{3D shape completion} %%% begin
% Shape completion is essentially an underconstrained problem.
Many traditional shape completion algorithms rely on geometric priors such as
volume smoothness to fill holes.  Poisson surface reconstruction~\cite{Kazhdan2006,Kazhdan2013} is one of the
representative methods. 
A few methods \cite{Sharf2004,Harary2014,Shen2012,Shao2012,Sung2015} fill the missing regions by synthesizing filling patches based on the geometry from the rest shape or a shape database. 
% However, the missing regions are usually needed to specify first.

% deep learning related
Neural networks have been extensively used for shape completion. 
In general, voxel-based networks~\cite{Wu2015,Wu2016}, Octree-based and kd-tree networks~\cite{Wang2017,Wang2018a,Hane2017,Riegler2017a,Tatarchenko2017,Klokov2017}, Point-based networks \cite{Qi2017a,Qi2017,Su2017},implicit function-based networks~\cite{Park2019,Chen2019,Mescheder2019} are all suitable to be adapted for this task.
Particularly,  Dai \etal~\shortcite{Dai2017} propose a 3D-Encoder-Predictor network,
which uses a voxel-based encoder-decoder network with skip connections to
regress the missing geometry.
%  Then the predicted shape is used
% for retrieving high-resolution shapes from the training set, with which the geometry
% details are refined via a patch-based 3D shape synthesis process.
Han \etal~\shortcite{Han2017}  propose to decompose the shape completion task
into two steps: global shape inference and local geometry refinement.
% Each step is accomplished with a dedicated voxel-based network. In their second step, the hole regions are assumed known, which is not very practical. 
Cao \etal~\shortcite{Cao2018} cascade OctNet-based fully convolutional sub-networks~\cite{Riegler2017} infer missing surface areas. %  and refine geometric details as a TSDF volume in a progressive, coarse-to-fine manner.
Other existing works focus either on reducing the full supervision to weakly-supervision~\shortcite{Stutz2018}, employ the generative adversarial loss to improve the shape completion quality \cite{Wu2018,Yang2018},
or use implicit function-based shape representation for shape completion \cite{Park2019,Mescheder2019}.

%%%%%%%%%%%%%%%%%%%%%%%%%%%%%%%%%%%%%%%%%%%%%%%%%%%%
\myskip\paragraph{3D scene semantic completion} %%% begin
% 3D scene semantic segmentation and completion are critical for 3D scene understanding
% and creation.
Based on the observation that scene semantic segmentation and completion are tightly
intertwined, Song \etal~\shortcite{Song2017} propose SSCNets to solve these two problems
simultaneously, which achieve superior performance than previous approaches
~\cite{Zheng2013,Firman2016}.
% However, SSCNets are based on 3D volumetric CNNs whose memory and computational cost are quite high, so they cannot support high-resolution 3D input and output, and cannot enjoy a very deep network architecture for improving the performance.
VVNets~\cite{Guo2018} combine 2D view-based CNNs and 3D volumetric CNNs, thus greatly
reduce the training and inference cost. 
% With deeper network architectures on the image domain, VVNets achieve much better
% performance than SSCNets.
By utilizing sparse convolution~\cite{Graham2017}, SGCNets~\cite{Zhang2018}  further improve the efficiency and performance for scene semantic completion.
% However, their decoder is still based on volumetric CNNs, thus they share the same limitation of other volumetric CNN-based approaches. 
In our method, we combine the octree-based U-Net with ResNet blocks and the specially
designed skip connections with deep layers, enabling a simple and effective solution.

\myskip\paragraph{Skip connections in deep learning} %%% begin
% Skip connections play an important role in modern deep learning architectures.
In the U-Net structure~\cite{Ronneberger2015}, the features from the encoder are concatenated with
the features in the decoder via skip connections for merging the spatial information
from the encoder into the decoder directly.
% The effectiveness of U-Net is verified by extensive works
% such as \cite{Long2015,Badrina2017}.
ResNet~\cite{He2016,He2016b}, uses skip connections to add the features between two or three consecutive convolution
layers, which greatly eliminates the gradient explosion/vanishing problems.

Different from the 2D image domain, for 3D shape completion with sparse 3D representations like octrees, the spatial locations of the input and output points are different. 
To address this issue, we propose \emph{output-guided skip connections}: skip connections are added only where there are output features. 
The output-guided skip connection not only reduces the complexity of the network but also connects the essential input and output features. 
% Our experiments show that the proposed skip connections are effective and greatly enhance the completion results.

%%% end 
\section{Network design}
\label{sec:method}

\begin{figure}[b]
  \centering
  \includegraphics[width=0.85\linewidth]{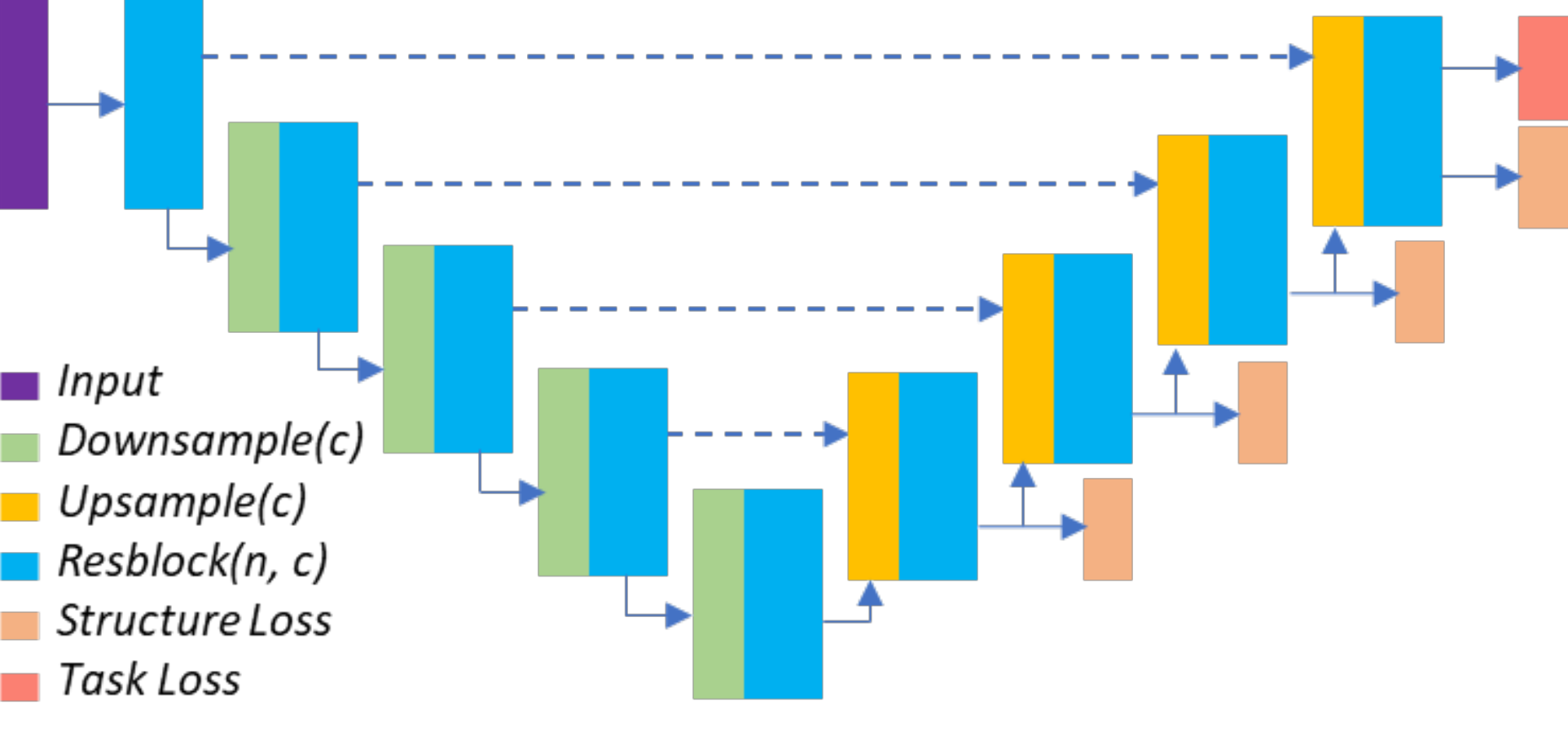}
  \caption{Our network architecture. $c$ is the channel number. The input and output are octrees.
   }
  \label{fig:network} %\vspace{-4mm}
\end{figure}

\begin{figure*}[t]
  \centering
  \includegraphics[width=0.85\linewidth]{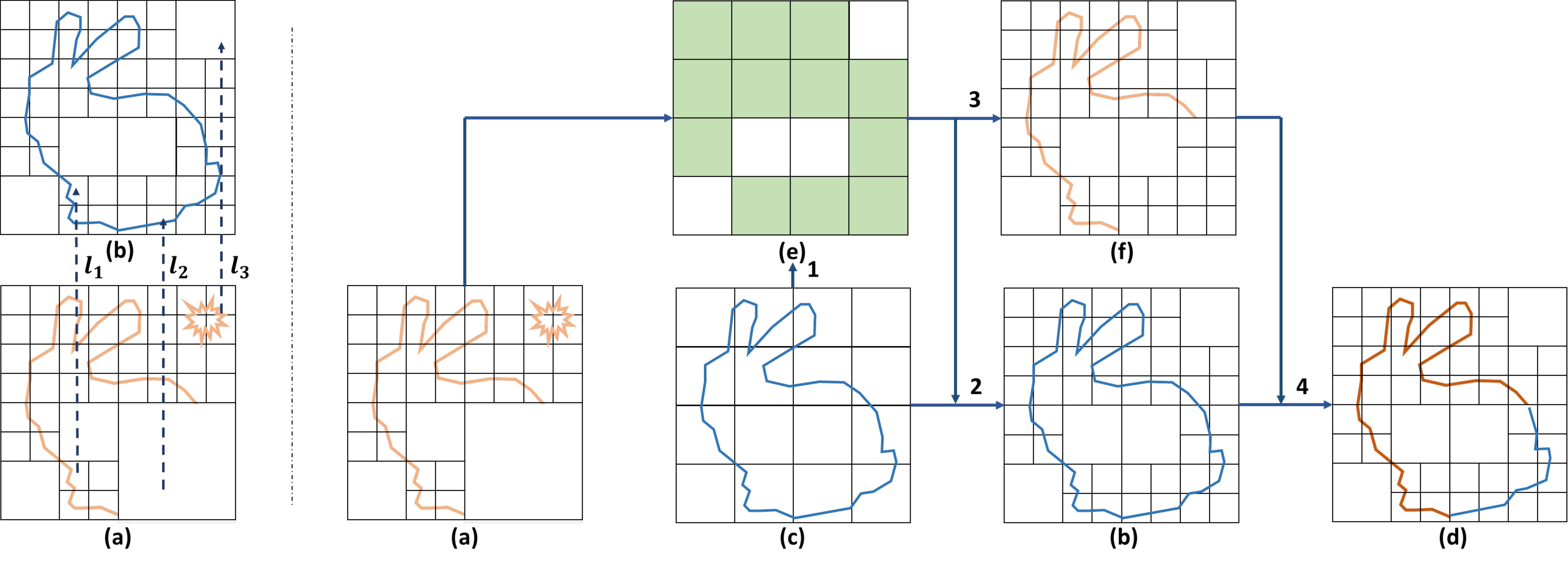} \vspace{-2mm}
  \caption{Output-guided skip connection.
  The left figures show three skip connections  $l_1$, $l_2$ and $l_3$ for analysis.
  Figures on the right show the construction of output-guided skip connections.
  %  (1) the prediction module of the decoder takes the features in (c) and predicts the
%  node status as shown in (e), where the green color means the node is non-empty;
%  (2) The octree in (c) is subdivided according to the status map (e);
%  (3) The features in (a) are multiplied with the mask of (e) and zeros are padded when the
%  corresponding nodes do not exist in (a), the result is in (f);
%  (4). Features in (f) and (b) are added together and result in the feature map of (d).
%  Here the dark yellow line in (d) is for highlighting that the spatial
%  information from (a) is added to (d).
 }
  \label{fig:link}  \vspace{-4mm}
\end{figure*}

\subsection{Network overview}

Our 3D completion network is built upon the octree-based autoencoder~\cite{Wang2018a}.
Multiple ResNet blocks~\cite{He2016} are stacked in the network. 
The encoder and decoder are linked via our output-guided skip connections.
The overall network architecture follows the U-Net design~\cite{Ronneberger2015},
as shown in Figure~\ref{fig:network}, the network details are present in Section~\ref{sub:netowrks}.

 % in which
 %  \emph{Downsample(c)} and \emph{Upsample(c)} are octree-based convolution and
 %  deconvolution operators with kernel size and stride as 2,
 %  \emph{Resblock(n, c)} is a stack of $n$ ResNet blocks,
 %  the dashed line represents the proposed skip connections,
 %  \emph{Structure Loss} is used to train the decoder for predicting the octree structure,
 %  and \emph{Task Loss} is used to regress the plane normal and displacement or output segmentation
 %  probability.

\myskip\paragraph{Input and output}
The network takes octrees built from the incomplete point cloud as the input and the ground-truth point cloud as the target output. The point cloud can be from 3D scans or other 3D forms that can be discretized as a point cloud, such as the voxelized shape.
We assume the point cloud is equipped with oriented normals, if not, we estimate normals from points.
% In our experiments, the depth of the octree is set to $6$, \ie the fine-level resolution is about $64^3$.

\myskip\paragraph{O-CNN encoder and decoder}
To make our paper self-explanatory, we briefly introduce the octree-based encoder
and decoder~\cite{Wang2018a}. 
The O-CNN encoder takes the octree as input and constrains the CNN computation within
the octree with the rule: \emph{where there is an octree node, there is CNN computation}.
After a series of octree-based CNN operations~\cite{Wang2017}, 
the generated feature maps are processed and down-sampled,
and flow along with the octree structure in a bottom-up manner.
With the O-CNN decoder, the target octree is generated in top-down order.
In each level of the octree, one shared prediction module (2-layer MLP) processes the
features contained in each octree node and predicts whether this octree node is
empty or not. If the node is predicted to be non-empty, it will be further subdivided, and the feature
on this node is passed to its children via an octree-based deconvolution operator.
This process is repeated recursively until the specified maximum octree depth is
reached. In the finest octree level, a local planar
patch is predicted at each non-empty octree node, \ie the plane normal and displacement, are
regressed as the final output.  
% The output shape can be assembled by these small planar patches contained in the
% finest octree nodes.

% In the following sections, we detail the proposed output-guided skip connections, network layers, and loss functions.

%%%%%%%%%%%%%%%%%%%%%%%%%%%%%%%%%%%%%%%%%%%%%%%%%%%%
\subsection{Output-guided  skip connections}  %%% begin
\label{sub:link}

% key observation
When using octree-based autoencoders for shape completion, the input and output octree structures are different. The input octree is constructed from a partial
shape which is even probably distorted by noise, while the desired output
octree contains the complete shape.
Due to this difference, the feature maps from the encoder are not aligned with
the decoder, thus cannot be directly added via skip connections.
We illustrate invalid alignments in Figure~\ref{fig:link}-left, where
The input partial shape (a)  contains additional noise on the top-right region and the ground-truth shape is in (b).
Here we use 2D shapes and their corresponding quadtrees for illustration.

We analyze three possible skip connections denoted by $l_1$, $l_2$, and $l_3$. We can see that $l_1$ is a valid skip connection since the corresponding input octree node contains information that needs to be retained in the output.
$l_2$ is useless when the features on the input node are zero. $l_3$ is undesired since the features from the noise region should not be passed to the decoder, otherwise, they would interfere with the prediction.

To address the above issues, we propose \emph{output-guided skip connections}. The basic idea is simple: the skip connection between the encoder and decoder is only added for a non-empty octree node in the output when there is an input octree node in the same location.

The output-guided skip connections are built as follows.
Denote the feature map of octree level $l$ in the decoder as $D_l$, the feature
vector of an octree node with integer coordinate $\mx=(x, y, z)$ as $D_l(\mx)$.
The shared prediction module in octree level $l$ takes $D_l(\mx)$ as input and
predicts whether the octree node is empty or not.
The output probability is rounded to 0 or 1 and denoted as $S_l(\mx)$.
The octree nodes with $S_l(\mx)=1$ are further subdivided, and
the coordinates of subdivided child octree nodes are $(2x+i, 2y+j, 2z+k)$
where $i, j, k \in \{0, 1\}$.
Similarly, denote the octree node feature of octree level $l$ in the encoder as
$E_l(\mx)$.
Then the proposed skip connections can be formally defined as:
\begin{equation}
  D_{l+1}(\mx) =
  D_{l+1}(\mx) +
    E_{l+1}(\mx) \cdot S_l(\mx/2).
  \label{equ:link}
\end{equation}
Here $E_{l+1}(\mx) = 0$ if there is no input octree node with coordinate $\mx$.
The arithmetic operation defined by Equation~\ref{equ:link} is applied to every
octree node and each channel of the feature map of the octree level $(l+1)$  in
the decoder respectively.
It is possible to use the unrounded version of $S_l$ in \eqref{equ:link} in our network. 
Experiments show the benefit is margin, so we always use the rounded version for simplicity.

We illustrate the output-guided skip connection in Figure~\ref{fig:link}-right in which where one feature map of the encoder (a) is added to one feature map of the decoder (b) and the result is shown as (d).
The overall operation includes 4 steps:
% \begin{enumerate}[leftmargin=*]\setlength\itemsep{0mm}
  \textbf{1.} The prediction module of the decoder takes the features in (c) and predicts the
  node status as shown in (e), where the green color means the node is non-empty, \ie $S_l(\mx)=1$;
  \textbf{2.} The octree in (c) is subdivided according to the status map (e);
  \textbf{3.} The features in (a) are multiplied with the mask of (e) and zeros are padded when the
  corresponding nodes do not exist in (a), the result is in (f);
  \textbf{4.} Features in (f) and (b) are added together and result in the feature map of (d).
  The dark yellow line in (d) is used for highlighting that the spatial
  information from (a) is added to (d).
 % \end{enumerate}
Note that the star shape, \ie the outlier in the noisy input (a) can be easily filtered by the output-guided skip connection,
which explains the robustness of our network to input noise conceptually.
% Experiments in Section~\ref{sub:completion} further verify this advantage.

% implementation details
The execution of output-guided skip connections is  very efficient.
The most expensive operation in Equation~\ref{equ:link} is to search the octree
node with the same coordinates for each octree node of the output octree.
As the coordinates are stored as \emph{shuffled keys} sorted in ascending order~\cite{Wilhelms1992,Zhou2011},
the searching operation can be executed efficiently via the parallel binary
search operator provided by the Thrust library~\cite{Kaczmarski2013}.

\myskip\paragraph{Remark}
OctNetFusion~\cite{Riegler2017a} proposes skip connections for
OctNet~\cite{Riegler2017} for the task of depth map fusion.
However, their skip connections are different from ours in nature.
OctNetFusion uses skip connections to increase the receptive field by 
\emph{statically} connecting feature maps from \emph{the same octree structure},
whereas our skip connections are used to constrain the network focusing on the 
predicted shape by \emph{dynamically} connecting feature maps from \emph{
the input and predicted octree structure}.
\looseness=-1

%%%%%%%%%%%%%%%%%%%%%%%%%%%%%%%%%%%%%%%%%%%%%%%%%%%%
\subsection{Network details}~\label{sub:netowrks}
The details of all the layers in Figure~\ref{fig:network} are as follows:
% \begin{enumerate}[leftmargin=*]\setlength\itemsep{0mm}
$\mathit{conv}(c, k, s)$ is the octree-based convolution 
followed by BN and ReLU,
where $c$ is the number of output channel, $k$ is the kernel size and $s$ is the stride.
$\mathit{Downsample}(c)$ is defined as $\mathit{conv(c, 2, 2)}$.
% , which means to downsample
% the feature map via a convolution operator with kernel size 2 and stride 2.
$\mathit{Resblock}(n, c)$ is a stack of $n$ ResNet blocks, each of which
is made up of  ``$\mathit{conv}(c/4, 3, 1) + \mathit{conv}(c, 3, 1)$''
with skip connections between them.
$\mathit{Upsample}(c)$: the octree-based deconvolution operator followed by
BN and ReLU. The kernel size and stride are set as 2, and the
output channel is $c$.
% \end{enumerate}
In our experiments, $c$ is set to 64 for the first $\mathit{Resblock}(n, c)$,
increases by a factor of 2 after encountering each $\mathit{Downsample}(c)$,
and decreases by a factor of 2 after every $\mathit{Upsample}(c)$.
The upper bound of $c$ is set to 256.

The prediction module of the decoder is a 2-layer MLP and outputs the probability of
octree node status between 0 and 1 with a Sigmoid function
% ($f(x) = 1/(1 + e^{-x}), x \in \mathbb{R}$) 
as the final activation function.
In the training stage, as the ground-truth octree node status is known, there is a
sigmoid cross-entropy loss in each level of the octree, which is called as
\emph{Structure Loss}: $L_{struct}^l$, $l$ is the level depth.
In the finest level of the octree, \emph{Task Loss}, denoted as $L_{task}$, are defined
for different tasks.
For shape completion (refer to Section~\ref{sub:completion}), % and~\ref{sub:skeleton}),
it has the following form:
$L_{task} = \frac{1}{n} \sum (\|\mathbf{n} - \mathbf{n}^*\|^2 + |d - d^*|^2)$,
where $(\mathbf{n}, d)$ and $(\mathbf{n}^*, d^*)$ are the predicted and ground-truth
planar parameters contained in the finest non-empty octree node,
respectively.
For semantic scene  completion (refer to Section~\ref{sub:scene}), the
network predicts the semantic labels of all non-empty voxels, so $L_{task}$
is the multi-class cross-entropy loss.
The total loss function of the network is defined as:
\begin{equation}
  Loss = \sum_{l = 3}^{d} L_{struct}^l + w \cdot L_{task},
  \label{equ:loss}
\end{equation}
where $d$ is the maximum octree depth, $w$ is a weight factor and set to 1 in our experiments.

%%% end 
\section{Experiments} \label{sec:result}

We evaluate our networks on the tasks of 3D shape completion and semantic scene completion.
% All the experiments were conducted on a desktop computer with one Intel(R) Core(TM)
% I7-5960K CPU (3.0GHz) and two GeForce GTX Titan X GPUs (12 GB memory).
% Our implementation is based on O-CNN framework~\cite{Wang2017} and 
We will release our code and models at \emph{https://github.com/microsoft/O-CNN}.

\begin{table}[t]
  \centering
  \vspace{2mm}
  \scalebox{0.74}{
  \begin{tabular}{l|cccccc}
  \toprule
  \footnotesize  & \footnotesize 3DEPN \cite{Dai2017} & \footnotesize 3DRecGAN \cite{Yang2018} & \footnotesize \emph{AE} &  \footnotesize \emph{Our\textsubscript{shallow}}   & \footnotesize \emph{Our\textsubscript{noise}}  & \footnotesize \emph{Our\textsubscript{deep}} \\ \midrule
  $D_c$(mean)    & 8.63  & 5.37 & 5.80        & 3.47           & 3.44        & \bf{3.06}      \\
  %$D_c$(SD)     & 7.49  & 6.23 & 6.89        & 4.62           & 4.74        & \bf{4.57}      \\
  \bottomrule
  \end{tabular}
  }
  \vspace{-2mm}
  \caption{The statistics of the mean Chamfer distances --$D_c$(mean) on the shape completion task. All the shapes are normalized and centered in a box with size $128$. }
  \label{tab:chamfer} \vspace{-3mm}
\end{table}

\begin{figure}
  \centering
  \begin{overpic}[width=0.9\linewidth]{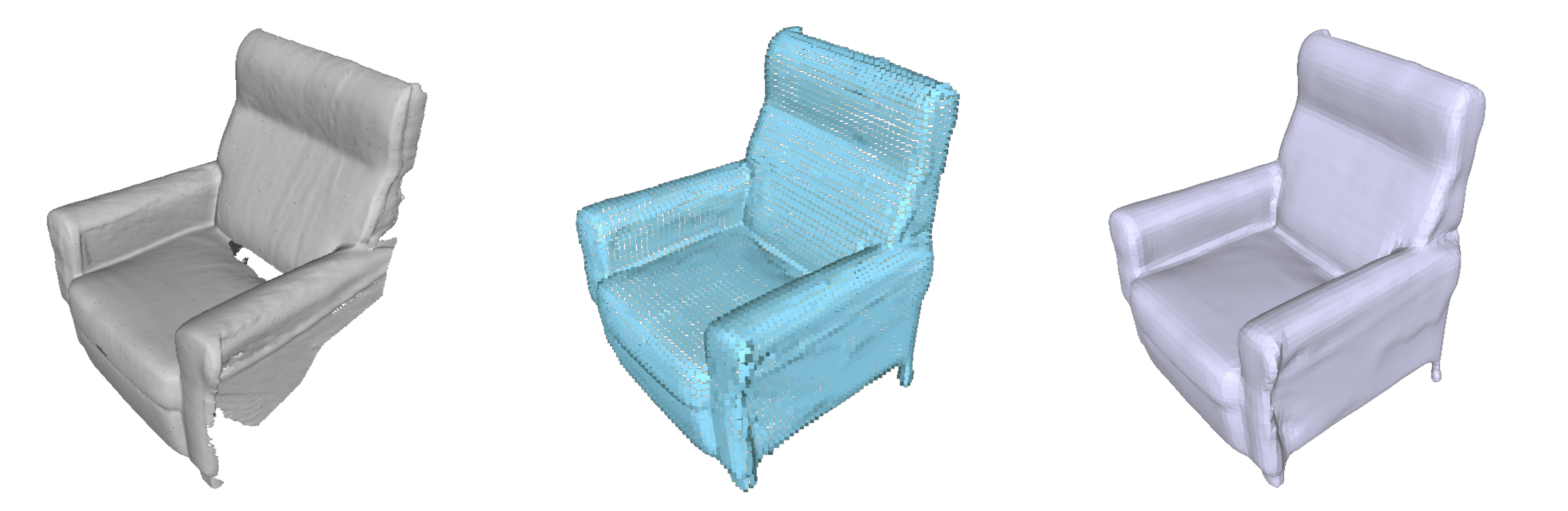}
    \put(10, -2){\footnotesize (a) Input}
    \put(42, -2){\footnotesize (b) Output}
    \put(73, -2){\footnotesize (c) Recons. Mesh}
  \end{overpic}
  % \vspace{-2mm}
  \caption{Shape completion on real data.
  The output points and reconstructed meshes are shown
  in column (b) and (c).
  % The input shapes (a) are scanned with a PrimeSense sensor.
  % The output point clouds of our network are shown in column (b).
  % The  reconstructed meshes are in column (c).
  }
  \label{fig:realdata}  \vspace{-4mm}
\end{figure}

%%%%%%%%%%%%%%%%%%%%%%%%%%%%%%%%%%%%%%%%%%%%%%%%%%%%
\subsection{3D shape completion} %%% begin
\label{sub:completion}
% We validate the efficacy of our method for 3D shape completion.
For 3D shape completion, the incomplete input is a point cloud from single or multiple registered 3D scans.
The goal is to fill the missing regions. % and reconstruct a complete 3D shape.

\myskip\paragraph{Dataset}
We use the dataset provided by~\cite{Dai2017}.
% , which is a subset of ShapeNet~\cite{Chang2015}.
There are 26790 3D objects from 8 categories,  25,590 objects for training and
1,200 objects for testing.
The partial scans are generated by virtual 3D scanning.
Each object has been scanned 1 to 6 times from different views.
% and 49\% of test data are single view scans.
These depth scans are back-projected into the original object space to form a
point cloud and each point is assigned with a normal which is estimated in from
depth scans.
We convert each incomplete point cloud to an octree of depth 6.
% , in which planar patch primitives are stored at the finest level of the octree~\cite{Wang2018a}.
% The depth of the octree is set as 6.
% as the geometric approximation
% error of the octree to the ground-truth surface is small enough for this task.

\begin{figure*}[pt]
  \centering
  \begin{overpic}[width=1\linewidth]{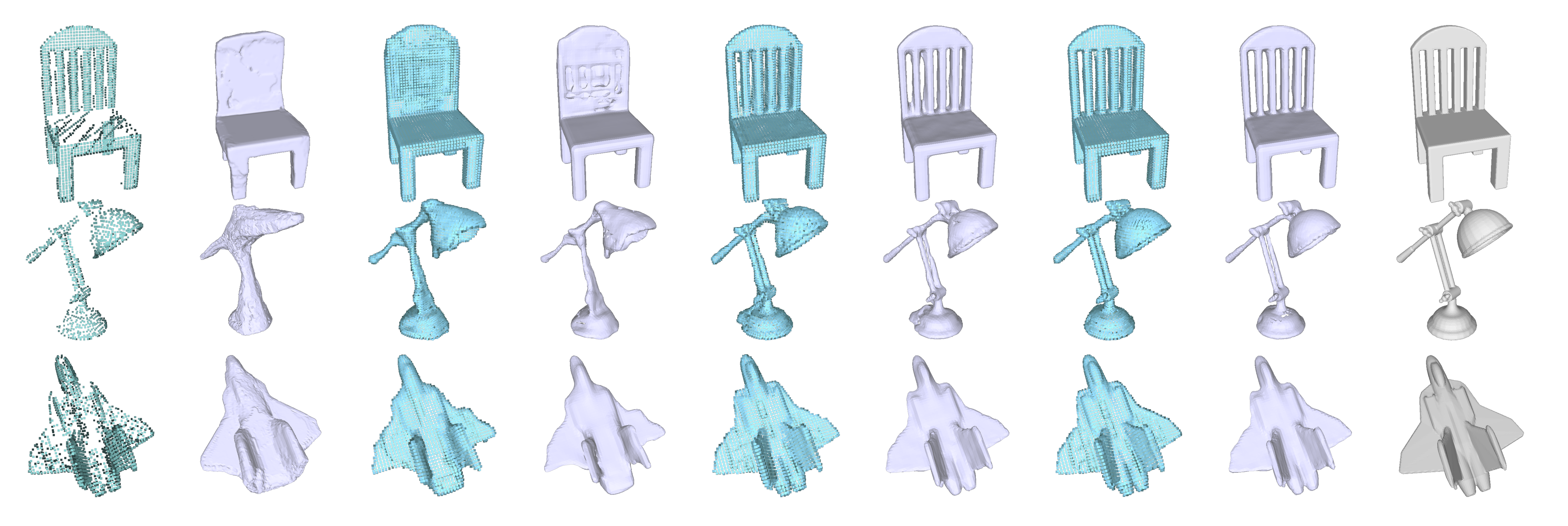}
    \put(3, -1.5){\small (a) \footnotesize Input}
    \put(14,-1.5){\small (b) \footnotesize 3DEPN}
    \put(25.5,-1.5){\small (c) \footnotesize \emph{AE}}
    \put(36.5,-1.5){\small (d) \footnotesize \emph{AE}}
    \put(44,-1.5){\small (e) \footnotesize \emph{Our\textsubscript{shallow}}}
    \put(56.5,-1.5){\small (f) \footnotesize \emph{Our\textsubscript{shallow}}}
    \put(68.5,-1.5){\small (g) \footnotesize \emph{Our\textsubscript{deep}}}
    \put(79,-1.5){\small (h) \footnotesize \emph{Our\textsubscript{deep}}}
    \put(90,-1.5){\small (i) \footnotesize Ground-truth}
    % \put(3, -1.5){\small (a) Input}
    % \put(14,-1.5){\small (b) 3DEPN}
    % \put(28,-1.5){\small (c) Autoencoder}
    % \put(50,-1.5){\small (d) Our(shallow)}
    % \put(74,-1.5){\small (e) Our(deep)}
    % \put(89,-1.5){\small (f) Ground-truth}
  \end{overpic}
  \vspace{-2mm}
  \caption{Visual comparison of single object completion.
  Figures in blue appearance are the \emph{raw point clouds} produced by our networks.
  % which are marked with a star symbol in the caption.
  Figures in gray appearance are the reconstructed meshes.
  Apparently, our results (g) and (h) are much more faithful to the ground-truth.}
  \label{fig:complete_vis1} % \vspace{-2mm}
  % \vspace{1mm}
\end{figure*}

\begin{figure*}
  \centering
  \begin{overpic}[width=1\linewidth]{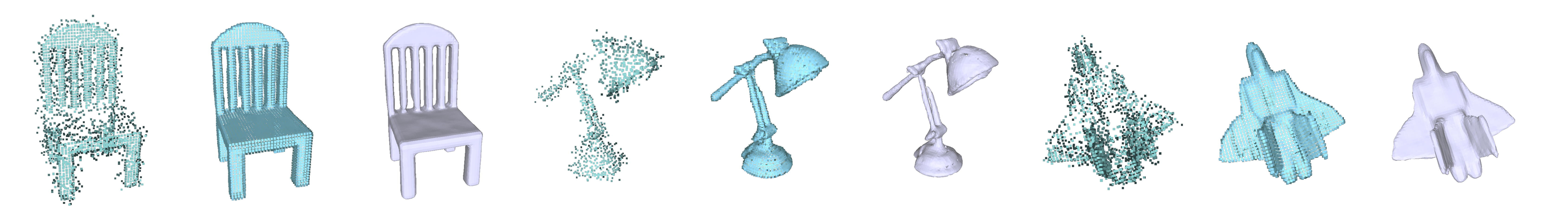}
    \put(3, -1){\small (a) \footnotesize Input}
    \put(12,-1){\small (b) \footnotesize \emph{Our\textsubscript{noise}}}
    \put(24,-1){\small (c) \footnotesize \emph{Our\textsubscript{noise}}}
    \put(37,-1){\small (d) \footnotesize Input}
    \put(47,-1){\small (e) \footnotesize \emph{Our\textsubscript{noise}}}
    \put(58,-1){\small (f) \footnotesize \emph{Our\textsubscript{noise}}}
    \put(70,-1){\small (g) \footnotesize Input}
    \put(79,-1){\small (h) \footnotesize \emph{Our\textsubscript{noise}}}
    \put(91,-1){\small (i) \footnotesize \emph{Our\textsubscript{noise}}}
    % \put(3, -1){\small (a) Input}
    % \put(17,-1){\small (b) Our(deep)}
    % \put(37,-1){\small (c) Input}
    % \put(52,-1){\small (d) Our(deep)}
    % \put(70,-1){\small (e) Input}
    % \put(85,-1){\small (f) Our(deep)}
  \end{overpic}
  \vspace{-2mm}
  \caption{Robustness test of our network. % with noisy input.
  The input noisy partial scans are in column (a), (d) and (g).
  The \emph{raw output point clouds} are in column (b), (f) and (h).
  The reconstructed meshes are in column (c), (f) and (i).
  The ground-truth meshes are the same as those in Figure~\ref{fig:complete_vis1}.}
  % With the noisy input, the quality of produced meshes is still better than 3DEPN. }
  \label{fig:complete_vis2}  \vspace{-4mm}
\end{figure*}

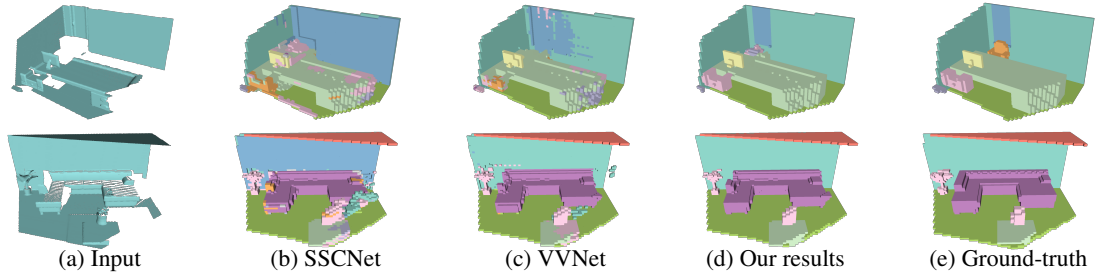
\begin{figure*}
  \centering
  \vspace{-2mm}  
  \begin{overpic}[width=0.85\linewidth]{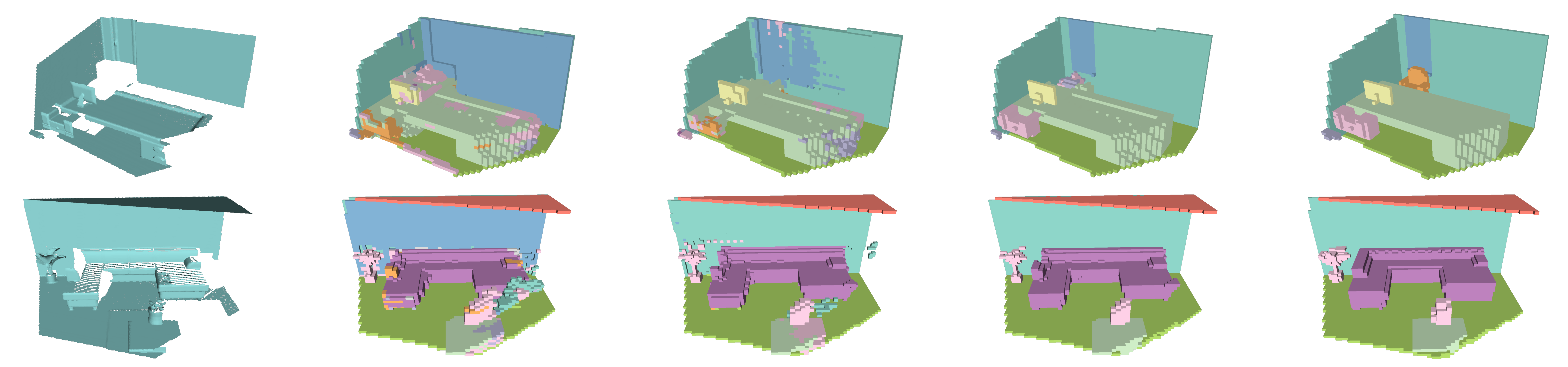}
    \put(6, -1){\small (a) Input}
    \put(25,-1){\small (b) SSCNet}
    \put(46,-1){\small (c) VVNet}
    \put(64,-1){\small (d) Our results}
    \put(84,-1){\small (e) Ground-truth}
  \end{overpic}
  % \vspace{-1mm}
  \caption{Visual results of semantic scene completion from one single depth image.
  Compared with SSCNet and VVNet, our results are much more faithful to the ground-truth.}
  \label{fig:scene}  \vspace{-2mm}
\end{figure*}

% iou table
\begin{table*}[t]
  \centering
    \scalebox{0.8}{
  \begin{tabular}{r|ccc|ccccccccccc|c}
  \toprule
          & \multicolumn{3}{c|}{Scene completion}     & \multicolumn{12}{c}{Semantic scene completion}                                   \\ \midrule
  Method  & prec.  & recall & IoU   & ceil.  & floor  & wall  & win.  & chair  & bed   & sofa   & table  & tvs   & furn.  & objs. & avg. \\ \midrule
  3DRecGAN~\cite{Yang2018} & -      & -      & 72.1  & 79.9   & 75.2   & 48.2  & 28.9  & 20.2   & 64.4  & 54.6   & 25.7   & 17.4  & 33.7   & 24.4  & 43.0 \\
  SSCNet~\cite{Song2017}    & 76.3   & 95.2   & 73.5  & 96.3   & 84.9   & 56.8  & 28.2  & 21.3   & 56.0  & 52.7   & 33.7   & 10.9  & 44.3   & 25.4  & 46.4 \\
  ForkNet~\cite{Wang2019}   & -      & -      & 86.9  & 95.0   & 85.9   & 73.2  & 54.5  & 46.0   & 81.3  & 74.2   & 42.8   & 31.9  & 63.1   & 49.3  & 63.4 \\
  SATNet~\cite{Liu2018a}    & 80.7   & 96.5   & 78.5  & 97.9   & 82.5   & 57.7  & 58.5  & 45.1   & 78.4  & 72.3   & 47.3   & 45.7  & 67.1   & 55.2  & 64.3 \\
  VVNet~\cite{Guo2018}      & 90.8   & 91.7   & 84.0  & 98.4   & 87.0   & 61.0  & 54.8  & 49.3   & 83.0  & 75.5   & 55.1   & 43.5  & 68.8   & 57.7  & 66.7 \\
  SGCNet~\cite{Zhang2018}   & 92.6   & 90.4   & 84.5  & 96.6   & 83.7   & 74.9  & 59.0  & 55.1   & 83.3  & 78.0   & 61.5   & 47.4  & 73.5   & 62.9  & 70.5 \\
  CCPNet~\cite{Zhang2019}   & 98.2   & 96.8   & 91.4  &\bf{99.2} &89.3 &76.2  &\bf{63.3} &58.2 &86.1 &\bf{82.6} &65.6 &53.2 &\bf{76.8} &65.2 &74.2\\ \midrule
  Our Results                      & 92.1   & 95.5   & 88.1  & 98.2   & \bf{92.8}   & \bf{76.3}  & 61.9  & \bf{62.4}   & \bf{87.5}  & 80.5   & \bf{66.3}   & \bf{55.2}  & 74.6   & \bf{67.8}  & \bf{74.8} \\
  \bottomrule
  \end{tabular}
  }
  \vspace{-2mm}
  \caption{ Quantitative comparison on the SUNCG dataset.
  The evaluation metric is Intersection over Union (IoU).
  % For scene completion, all non-empty voxels are treated as one category, and the
  % IoU is evaluated for the occluded voxels in the view frustum.
  % For semantic completion, the IoU for each class is evaluated on all the
  % non-empty voxels in the view frustum.
  Better results are in bold font.
  Our method outperforms other state-of-art methods on the average IoU.
  } 
  \vspace{-4mm}
  \label{tab:scene}
\end{table*}

\myskip\paragraph{Implementation details}
As our output is an octree in which each finest leaf node contains a planar patch,
we can sample multiple points on each patch and construct a point cloud.
% $S = \{\hat{x}_i\}_{i=1}^m$ for representing the predicated shape.
The approximation quality   to the ground-truth complete point cloud
is measured by using the discrete Chamfer distance metric.
% $D_c$:
% \begin{equation*}
%  D_c = \frac{1}{n} \sum_{x_i \in S_g} \min_{\hat{x}_j \in S} \| x_i - \hat{x}_j \|_2^2 +
%   \frac{1}{m} \sum_{\hat{x}_i \in S} \min_{x_j \in S_g} \| \hat{x}_i - x_j \|_2^2.
% \label{equ:chamfer}
% \end{equation*}
% \myskip\paragraph{Network and training details}
We set $n=2$ for each Resnet block, resulting in a deep network with 51 layers.
The training details are provided in supplementary materials.
% We set the batch size to 12, and stochastic
% gradient descent (SGD) with a momentum of 0.9 and a weight decay of 0.0005 is used for training.
% The initial learning rate is set as 0.1 and decreases by a factor of 10 after
% about 6 epochs. The training process stops after 25 epochs.
% % We  use two GPUs to train the network.

\myskip\paragraph{Comparison}
We compared our method with state-of-the-art methods: 3D-Encoder-Predictor
CNN (3DEPN)~\cite{Dai2017} and 3DRecGAN~\cite{Yang2018}.
% 3DEPN is a volumetric encoder-decoder network with skip connections added in
% the whole volume. 3DRecGAN improves 3DEPN via adding a GAN loss.
They take the partial TSDF~\cite{Curless1996} built from partial scans and regress the complete TSDF, with which the output mesh can be extracted.
% via the marching cube algorithm.
For evaluating the Chamfer distance, we sample dense points on the extracted mesh.
% As our octree is with depth 6, we subdivide the planar patch in each finest-level cuboid into 8 sub-cuboids and sample one point from each non-empty sub-cuboid.

As deep layers and output-guided skip connections are the key components of our network, we also design alternative networks to justify their importance. 
\vspace{-1mm}
\begin{enumerate}[leftmargin=*]\setlength\itemsep{0mm}
  \item[-] \emph{Shallow network}: \emph{Our\textsubscript{shallow}}.
  We use a shallow network (14 layers) with a similar amount of trainable parameters to 3DEPN.  The network is trained with the same training settings.
  \item[-] \emph{No skip connections}: \emph{AE}.
  By removing the proposed skip connections, the network is essentially an octree-based autoencoder~\cite{Wang2018a} with 51 layers.
  \end{enumerate}
\vspace{-1mm}
The mean Chamfer distances of all the methods are summarized in Table~\ref{tab:chamfer}.
% , and visual comparisons on some models can be found in ~\ref{fig:complete_vis1}.
We observe the following facts: % from the statistics.
\vspace{-1mm}
\begin{enumerate}[leftmargin=*]\setlength\itemsep{0mm}
  \item[-] Our deep and shallow network with output-guided skip connection outperform 3DEPN and 3DRecGAN significantly, which proves the superiority of combining octree with our proposed skip connections over volumetric TSDF with original skip connections everywhere. We explain this superiority is because our network constrains the CNN computation around the predicted surface and puts more focus on the predicted shape via output-guided skip connection, compared with 3DEPN and 3DRecGAN which have to predict all the voxels with high cost.
  \item[-] Without skip connection, the performance of our deep network drops and is even worse than our shallow network with the skip connection.  The completion results tend to be blurry and the geometric features are lost in some detailed regions as can be found in Figure~\ref{fig:complete_vis1}. The result verifies that the spatial information contained in the input is essential and skip connection can well communicate this information for the completion task.
  \end{enumerate}
\vspace{-3mm}

% noisy data
% \vspace{-2mm}
\myskip\paragraph{Robustness}
To verify the robustness of our network, we added Gaussian noise to the depth scan in the training dataset
and train the network again with the same training settings.
The mean of the Gaussian is set as 0, and the standard deviation is set as 2.5\%
of the width of the original object bounding box. 
We denote this training network by ${Our}_{noise}$.
From Table~\ref{tab:chamfer}, we can see that the performance drops slightly compared with our deep network trained on clean data, but clearly better than other methods. 
% This experiment well demonstrates our network is robust to noise, due to the use of output-oriented skip connection which helps to filter noise, as explained early.

% visual results
\myskip\paragraph{Result visualization}
We uniformly sample points with normals on the predicted octree and reconstruct
the meshes via Poisson Reconstruction~\cite{Kazhdan2013}.
And the results are shown in Figure~\ref{fig:complete_vis1}. 
The input partial scans and the ground-truth meshes  are shown in column (a) and column (g).

It is clear to see that the geometric fidelity of results from our shallow network result (f) is much better than 3DEPN (b). With the deep network, the results are further enhanced and close to the ground-truth. 
As the deep autoencoder does not utilize the skip connection, its output quality is even worse than our shallow network, despite using deep layers.

We illustrate the completion results from noisy partial inputs in Figure~\ref{fig:complete_vis2}.
It can be seen that the output quality is high, and better than 3DEPN with clean partial scans and our AE, which verifies the robustness of our method.

%%% todo: move this paragraph to supplementary materials
%Wang \etal~\cite{Wang2018a} used the Adaptive O-CNN based autoencoder for shape completion.
%The network was trained on the car dataset from the ShapeNet.
%We apply our network to one of the cars, and the result is shown in Figure~\ref{fig:aocnn}.
%Although the complete shape  of Adaptive O-CNN (c) looks reasonable, the spatial
%information contained in the incomplete shape is not fully preserved due to the
%absence of the skip connections.
%For example, the geometry structure marked by the red box is smoothed out
%compared with our result (d).

% real data
% \myskip\paragraph{Real data}
We also tried our trained network on real scans as shown in Figure~\ref{fig:realdata}.
The real data is provided by~\cite{Qi2016a}, which is scanned with a PrimeSense
sensor. % and reconstructed via the VoxelHashing method~\cite{Niessner2013}.
It can be seen our completion results are plausible.

%%% todo: move this figure to supplementary materials
%\begin{figure}[t]
%  \centering
%  \begin{overpic}[width=0.9\linewidth]{aocnn}
%    \put(18, 25){\footnotesize (a) Input}
%    \put(65, 25){\footnotesize (b) Ground-truth}
%    \put(10, -1.5){\footnotesize (c) Adaptive O-CNN}
%    \put(67, -1.5){\footnotesize (d) Our result}
%  \end{overpic}
%  \vspace{2mm}
%  \caption{Comparison with Adaptive O-CNN.
%  The incomplete point cloud is (a), the ground-truth is (b), the  output of Adaptive
%  O-CNN is (c) and our result is (d).
%  As highlighted by the red box, the geometry feature of our result is well preserved.
%  }
%  \label{fig:aocnn}  \vspace{-4mm}
%\end{figure}

% \begin{figure*}
%   \centering
%   \begin{overpic}[width=0.92\linewidth]{skeleton_vis}
%     \put(6,-1){\small (a) Input}
%     \put(25,-1){\small (b) P2P-Net}
%     \put(44,-1){\small (c) Our results}
%     \put(64,-1){\small (d) Recons. mesh}
%     \put(84,-1){\small (e) Ground-truth}
%   \end{overpic}
%   \caption{Visual results of shape reconstruction from meso-skeletons.
%   Compared with P2P-Net, the point cloud of our method is regularly distributed.
%   And since the normal is also regressed, the complete meshes can be reconstructed,
%   which is hard or even impossible for the point clouds of P2P-Net.}
%   \label{fig:skeleton}
% \end{figure*}

%%% end

%%%%%%%%%%%%%%%%%%%%%%%%%%%%%%%%%%%%%%%%%%%%%%%%%%%%
\subsection{Semantic scene completion from a depth image} %%% begin
\label{sub:scene}

% problem definition
The goal %of semantic scene completion from a single depth image 
is to predict the
occupancy and semantic labels in the view frustum for single depth
images of indoor scenes.

\myskip\paragraph{Dataset}
We use the SUNCG dataset provided by~\cite{Song2017}.% for the training and testing.
The training/testing dataset contains 150k/470 depth images and the corresponding
ground-truth label volumes.
% , which are generated by virtually scanning from 8k scenes.
% The testing dataset contains 470 depth images and the corresponding ground-truth
% label volumes, which are generated from another 170 scenes. %, which are not in the training set.
We convert the depth images to point clouds with normals and build octrees with depth 8.
The resolution of the ground-truth volumes is $60 \times 36 \times 60$, we convert
the non-empty voxels to point clouds with labels, and build  octrees with depth
6.

\myskip\paragraph{Implementation details}
We use the intersection over union (IoU) between the predicted voxels and the
ground-truth voxels as the evaluation metric.
% For the semantic completion task, the IoU for each class is evaluated on all the
% non-empty voxels in the view frustum.
% For the scene completion task, all non-empty voxels from different categories are
% treated as one category, and the IoU is evaluated for the occluded voxels in the
% view frustum.
% \myskip\paragraph{Network and training details}
In our network, we set $n=3$ for $Resblock(n,c)$.
Since the network in Figure~\ref{fig:network} takes the octree of depth 6 as input
while the depth of octree is 8 in this experiment, we add the following O-CNN blocks
to process and downsample the signal:
\vspace{-1mm}
\begin{equation*}
\begin{aligned}
  input& \rightarrow conv(3,1,16) \rightarrow pooling \rightarrow conv(3, 1, 16) \\
         &\rightarrow Resblock(32,1) \rightarrow conv(2,2,64)
\end{aligned}
\end{equation*}
% \vspace{-1mm}
In total, the network layer depth is 72.
The training details are provided in supplementary materials.
% The batch size is set to 8, and the network is trained with SGD with a
% momentum of 0.9 and a weight decay of 0.0005.
% The initial learning rate is set to 0.1 and decreases by a factor of 10 after
% about 6 epochs. The training process stops after 25 epochs.
% % We use two GPUs to train the network.

\myskip\paragraph{Comparison}
We validate the effectiveness of our method and compare it with state-of-the-art methods: 3DRecGAN~\cite{Yang2018}, ForkNet~\cite{Wang2019}, SSCNet~\cite{Song2017}, SATNet~\cite{Liu2018a}, VVNet~\cite{Guo2018},
SGCNet~\cite{Zhang2018} and CCPNet~\cite{Zhang2019}.  
Among them,  SGCNet~\cite{Zhang2019} shares some similarities with our network, which is also based on U-Net and uses the sparse convolution~\cite{Graham2017} in the encoder. However, SGCNet's decoder is based on volumetric CNNs.

% Notably, the training protocol we used in this experiment is almost the same as
% the one we used in the shape completion except for the batch size.
For simplicity, we did not balance the training data as
 \cite{Song2017,Zhang2018,Zhang2019} have done
or use the average voting trick as \cite{Guo2018,Zhang2019} have used, 
although these tricks are known to improve the network performance.
The evaluation results are summarized in Table~\ref{tab:scene}.
Our method achieves the best results on the average IoU metric in semantic scene completion.

We did a simple ablation study on this task.
First, we use 2 Resblocks and reduce the network depth to 54 (with a similar amount of
parameters and network depth to SGCNet),
the average IoU of semantic scene completion drops from 74.2\% to 70.9\%;
Second, we train the network without the output-guided skip connections, and the
IoU drops from 74.8\% to 49.3\%.
 The ablation study proves the importance of using deep layers and output-guided skip connections.

\myskip\paragraph{Visual results}
The complete scenes by our method are illustrated in Figure~\ref{fig:scene}.
The output is in the voxelized representation and the colors represent
different semantic labels. Here we also compare the results of VVNet and SSCNet whose implementation is available to the public.
Our results are clearly much more faithful to the ground-truth than the competitive methods.

\section{Conclusion} \label{sec:conclusion}
We proposed simple yet effective octree-based networks for shape and scene completion. % from noisy and incomplete inputs.
Our network achieves significant improvements in prediction accuracy, with the aid of our output-guided skip connections and the very deep octree-based network structures.  Experiments well demonstrate that our network outperforms the state-of-the-art work.

\newcount\cvprrulercount

\appendix

%%%%%%%%%%%%%%%%%%%%%%%%%%%%%%%%%%%%%%%%%%%%%%%%%%%%%%%%%%%%%%%%%%%

\section{Acknowledgements}
We wish to thank the anonymous reviewers for their constructive feedback, 
Chun-Yu Sun and Yu-Xiao Guo for preparing the dataset.

%%%%%%%%%%%%%%%%%%%%%%%%%%%%%%%%%%%%%%%%%%%%%%%%%%%%%%%%%%%%%%%%%%%

\section{Discussions}

\paragraph{Comparison to implicit function-based approaches}
Recently, implicit functions are used in deep learning as the 3D shape representation~\cite{Chen2019,Mescheder2019,Park2019}.
We did not conduct experiments to compare with these methods since they are 
\emph{orthogonal} to our method: 
they focus on the shape representation whereas we focus on network structures.
Technically, DeepSDF~\cite{Park2019} was applied to shape completion by optimizing
the latent code to match the partial data while completing the missing part, per shape, 
in a computationally expensive and memory-costly way. 
Our method can directly output the shape in one single forward pass.
OccNet~\cite{Mescheder2019} use the autoencoder architecture directly 
without skip-connection, the partial input cannot be well preserved. 
IM-Net~\cite{Chen2019} has not been tested the completion task.

\paragraph{Ablation study on skip connections $l_2$ and $l_3$ in Figure 2}
3DEPN is based on a dense U-Net and the decoder of SGCNet is also a dense network.
We regard them as comparable dense networks and do the comparisons with similar 
amount of parameters and network depth  in Section 4.
3DEPN and SGCNet use full skip-connections, including $l_2$ and $l_3$. 
In the ablation studies, our network without $l_2$ and $l_3$  achieves better results.

\section{Training details for the experiments}
For the experiments in Section 4.1 and 4.2, we use the same set of training parameters.
Specifically, we set the batch size to 8 and the weight decay to 0.0005, use stochastic
gradient descent (SGD) with a momentum of 0.9.
The initial learning rate is set as 0.1 and decreases by a factor of 10 after
about 6 epochs. The training process stops after 25 epochs.

%%%%%%%%%%%%%%%%%%%%%%%%%%%%%%%%%%%%%%%%%%%%%%%%%%%
\section{Shape reconstruction from a meso-skeleton} %%% begin
% \label{sub:skeleton}

To demonstrate the flexibility of the proposed method, we also conduct experiment 
on the task of reconstructing a complete 3D shape from its meso-skeleton.
\begin{figure*}
  \centering
  \begin{overpic}[width=0.9\linewidth]{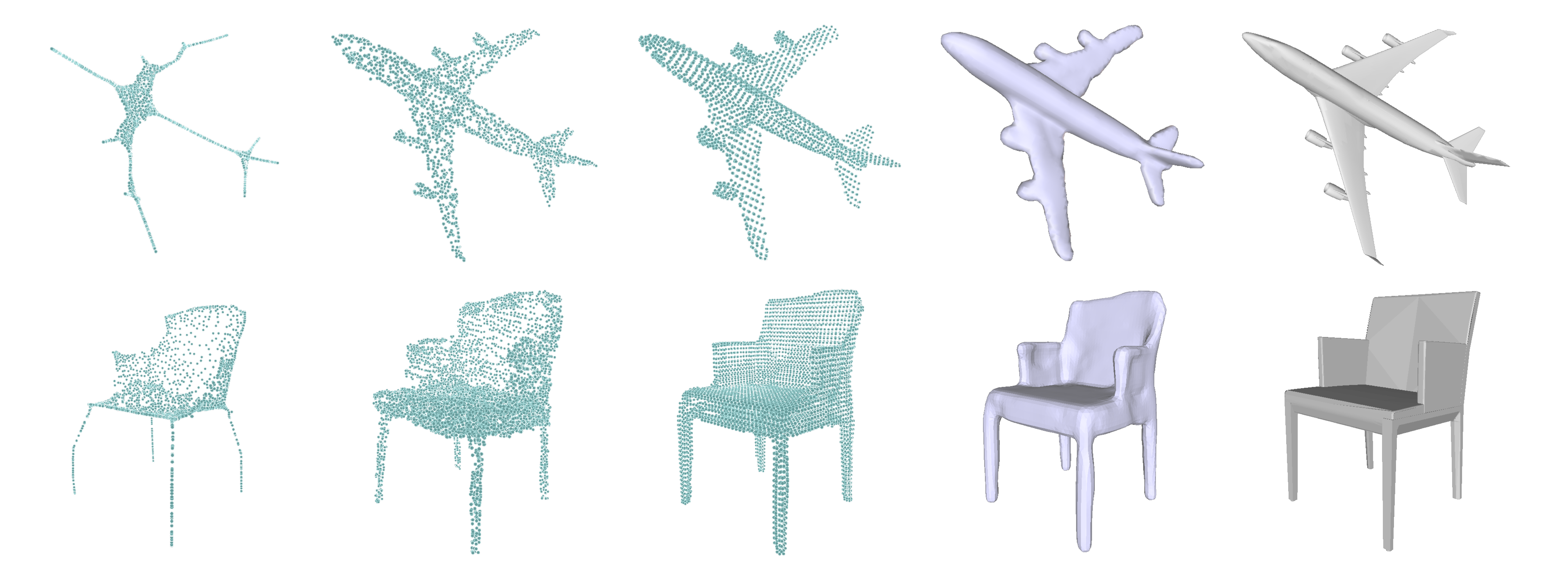}
    \put(6,-1){\small (a) Input}
    \put(25,-1){\small (b) P2P-Net}
    \put(44,-1){\small (c) Our results}
    \put(64,-1){\small (d) Recons. mesh}
    \put(84,-1){\small (e) Ground-truth}
  \end{overpic}
  \caption{Visual results of shape reconstruction from meso-skeletons.
  Compared with P2P-Net, the point cloud of our method is regularly distributed.
  And since the normal is also regressed, the complete meshes can be reconstructed,
  which is hard or even impossible for the point clouds of P2P-Net.}
  \label{fig:skeleton}
  \vspace{-2mm}
\end{figure*}
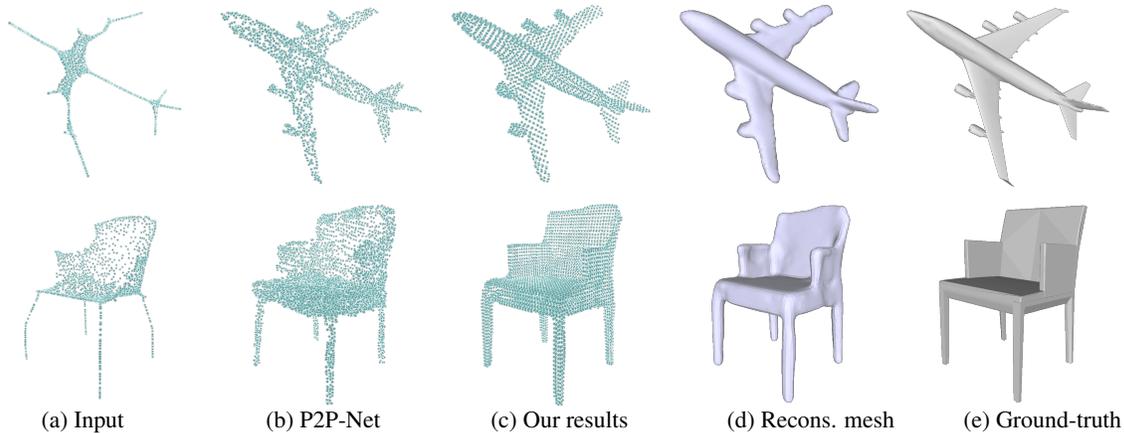

\vspace{-2mm}
\paragraph{Dataset}
We use the chair and plane datasets provided by~\cite{Yin2018}, which
include the synthesized meso-skeletons and the corresponding 3D shapes.
% These datasets originally come from the ModelNet40~\cite{Wu2015}.
The chair dataset contains 889 training and 100 testing pairs, and the
airplane dataset contains 626 training and 100 testing pairs.
The meso-skeletons are represented as point cloud containing 2048 points, with
which we build the octree directly.
For the 3D shapes, we use the virtual scanner to convert them into dense point
cloud with oriented normals~\cite{Wang2017}, then build the target octrees.
The depth of octree is set as 6.
The two datasets are trained \emph{separately}, which is the same as P2P-NET.

\vspace{-2mm}
\paragraph{Implementation details}
We use the same network as the one used in shape completion.
To avoid overfitting, we rotate each skeleton and the corresponding ground-truth
object along with the upright axis 12 times for data augmentation.
The batch size is set as 24, and the network is trained using SGD with a
momentum of 0.9 and a weight decay of 0.0005.
The initial learning rate is set as 0.1 and decreases by a factor of 2 after
60 epochs.
The training process stops after 120 epochs.
We  use the Chamfer distance defined in as the evaluation metric.

\vspace{-2mm}
\paragraph{Comparison}
We do a comparison with P2P-Net~\cite{Yin2018}.
Since there is no explicit point correspondence between the meso-skeleton and
the target shape, P2P-Net relies on a loss function enforcing a shape-wise
similarity between the predicted and the target point sets during the training
stage to build the correspondence.
We directly build the correspondence between the input meso-skeleton and
output shape with the proposed skip connections.
On the dataset plane and chair, the medians of Chamfer distances  are
1.04 and 5.55 for our methods, 1.66 and 6.06 for P2P-Net.

\vspace{-2mm}
\paragraph{Visual results}
The visual results are shown in Figure~\ref{fig:skeleton}.
Compared with P2P-Net, the point clouds produced by our method are regularly distributed.
Since the point normal is also regressed with the loss function, 
the output point cloud can be directly used as the input
of the Poisson Reconstruction method.
The reconstructed meshes are shown in the fourth column of Figure~\ref{fig:skeleton}.
However, it is very hard and even impossible to reconstruct surface mesh from the
point cloud of P2P-Net, since the point cloud of P2P-Net is scattered and the internal
volume structure of the shape is not kept, which makes it extremely difficult to
define the inside and outside for the shape.

%%% end 

{\small
  \bibliographystyle{ieee_fullname}
  \bibliography{src/reference}

\begin{thebibliography}{10}\itemsep=-1pt

\bibitem{Calakli2011}
F. Calakli and G. Taubin.
\newblock {SSD: Smooth signed distance surface reconstruction}.
\newblock {\em Comput. Graph. Forum}, 30(7), 2011.

\bibitem{Cao2018}
Yan-Pei Cao, Zheng-Ning Liu, Zheng-Fei Kuang, Leif Kobbelt, and Shi-Min Hu.
\newblock Learning to reconstruct high-quality {3D} shapes with cascaded fully
  convolutional networks.
\newblock In {\em European Conference on Computer Vision (ECCV)}, 2018.

\bibitem{Chen2019}
Zhiqin Chen and Hao Zhang.
\newblock Learning implicit fields for generative shape modeling.
\newblock In {\em Computer Vision and Pattern Recognition (CVPR)}, 2019.

\bibitem{Curless1996}
Brian Curless and Marc Levoy.
\newblock {A Volumetric Method for Building Complex Models from Range Images}.
\newblock In {\em SIGGRAPH}, 1996.

\bibitem{Dai2017}
Angela Dai, Charles~Ruizhongtai Qi, and Matthias Niessner.
\newblock {Shape completion using 3D-encoder-predictor CNNs and shape
  synthesis}.
\newblock In {\em Computer Vision and Pattern Recognition (CVPR)}, 2017.

\bibitem{Dai2018}
Angela Dai, Daniel Ritchie, Martin Bokeloh, Scott Reed, J{\"u}rgen Sturm, and
  Matthias Nie{\ss}ner.
\newblock {{ScanComplete}: Large-Scale Scene Completion and Semantic
  Segmentation for 3D Scans}.
\newblock In {\em Computer Vision and Pattern Recognition (CVPR)}, 2018.

\bibitem{Firman2016}
Michael Firman, Oisin~Mac Aodha, Simon Julier, and Gabriel~J. Brostow.
\newblock {Structured prediction of unobserved voxels from a single depth
  image}.
\newblock In {\em Computer Vision and Pattern Recognition (CVPR)}, 2016.

\bibitem{Goodfellow2016}
Ian Goodfellow, Yoshua Bengio, and Aaron Courville.
\newblock {\em {Deep Learning}}.
\newblock MIT Press, 2016.

\bibitem{Graham2017}
Benjamin Graham and Laurens van~der Maaten.
\newblock {Submanifold Sparse Convolutional Networks}.
\newblock {\em arXiv preprint arXiv:1706.01307}, 2017.

\bibitem{Guo2018}
Yuxiao Guo and Xin Tong.
\newblock View-volume network for semantic scene completion from a single depth
  image.
\newblock In {\em IJCAI}, 2018.

\bibitem{Han2017}
Xiaoguang Han, Zhen Li, Haibin Huang, Evangelos Kalogerakis, and Yizhou Yu.
\newblock {High-resolution shape completion using deep neural networks for
  global structure and local geometry inference}.
\newblock In {\em International Conference on Computer Vision (ICCV)}, 2017.

\bibitem{Hane2017}
Christian H{\"a}ne, Shubham Tulsiani, and Jitendra Malik.
\newblock {Hierarchical surface prediction for 3D object reconstruction}.
\newblock In {\em Proc. Int. Conf. on 3D Vision (3DV)}, 2017.

\bibitem{Harary2014}
Gur Harary, Ayellet Tal, and Eitan Grinspun.
\newblock {Context-based Coherent Surface Completion}.
\newblock {\em ACM Trans. Graph.}, 33(1), 2014.

\bibitem{He2016}
K. He, X. Zhang, S. Ren, and J. Sun.
\newblock {Deep residual learning for image recognition}.
\newblock In {\em Computer Vision and Pattern Recognition (CVPR)}, 2016.

\bibitem{He2016b}
Kaiming He, Xiangyu Zhang, Shaoqing Ren, and Jian Sun.
\newblock {Identity mappings in deep residual networks}.
\newblock In {\em European Conference on Computer Vision (ECCV)}, 2016.

\bibitem{Kaczmarski2013}
Krzysztof Kaczmarski and Pawel Rzazewski.
\newblock {Thrust and CUDA in data intensive algorithms}.
\newblock In {\em New Trends in Databases and Information Systems}, 2013.

\bibitem{Kazhdan2006}
Michael Kazhdan, Matthew Bolitho, and Hugues Hoppe.
\newblock {Poisson surface reconstruction}.
\newblock In {\em Symp. Geom. Proc.} Eurographics Association, 2006.

\bibitem{Kazhdan2013}
Michael Kazhdan and Hugues Hoppe.
\newblock {Screened Poisson surface reconstruction}.
\newblock {\em ACM Trans. Graph.}, 32(3), 2013.

\bibitem{Klokov2017}
Roman Klokov and Victor Lempitsky.
\newblock {Escape from cells: Deep Kd-networks for the recognition of 3D point
  cloud models}.
\newblock In {\em International Conference on Computer Vision (ICCV)}, 2017.

\bibitem{Liu2018a}
Shice Liu, Yu Hu, Yiming Zeng, Qiankun Tang, Beibei Jin, Yinhe Han, and Xiaowei
  Li.
\newblock See and think: Disentangling semantic scene completion.
\newblock In {\em Neural Information Processing Systems (NeurIPS)}, 2018.

\bibitem{Mescheder2019}
Lars Mescheder, Michael Oechsle, Michael Niemeyer, Sebastian Nowozin, and
  Andreas Geiger.
\newblock Occupancy networks: Learning 3d reconstruction in function space.
\newblock In {\em Computer Vision and Pattern Recognition (CVPR)}, 2019.

\bibitem{Park2019}
Jeong~Joon Park, Peter Florence, Julian Straub, Richard Newcombe, and Steven
  Lovegrove.
\newblock {DeepSDF}: Learning continuous signed distance functions for shape
  representation.
\newblock In {\em Computer Vision and Pattern Recognition (CVPR)}, 2019.

\bibitem{Qi2017a}
Charles~R. Qi, Hao Su, Kaichun Mo, and Leonidas~J. Guibas.
\newblock { PointNet: Deep learning on point sets for 3D classification and
  segmentation}.
\newblock In {\em Computer Vision and Pattern Recognition (CVPR)}. IEEE, 2017.

\bibitem{Qi2016a}
Charles~Ruizhongtai Qi, Hao Su, Matthias Nie{\ss}ner, Angela Dai, Mengyuan Yan,
  and Leonidas~J. Guibas.
\newblock {Volumetric and multi-view CNNs for object classification on 3D
  data}.
\newblock In {\em Computer Vision and Pattern Recognition (CVPR)}, 2016.

\bibitem{Qi2017}
Charles~R Qi, Li Yi, Hao Su, and Leonidas~J Guibas.
\newblock {PointNet++: Deep hierarchical feature learning on point sets in a
  metric space}.
\newblock In {\em Neural Information Processing Systems (NeurIPS)}, 2017.

\bibitem{Riegler2017a}
Gernot Riegler, Ali~Osman Ulusoy, Horst Bischof, and Andreas Geiger.
\newblock {OctNetFusion: Learning depth fusion from data}.
\newblock In {\em Proc. Int. Conf. on 3D Vision (3DV)}, 2017.

\bibitem{Riegler2017}
Gernot Riegler, Ali~Osman Ulusoy, and Andreas Geiger.
\newblock {OctNet: Learning deep 3D representations at high resolutions}.
\newblock In {\em Computer Vision and Pattern Recognition (CVPR)}. IEEE, 2017.

\bibitem{Ronneberger2015}
Olaf Ronneberger, Philipp Fischer, and Thomas Brox.
\newblock {{U-Net}: Convolutional networks for biomedical image segmentation}.
\newblock In {\em International Conference on Medical image computing and
  computer-assisted intervention}, pages 234--241, 2015.

\bibitem{Shao2012}
Tianjia Shao, Weiwei Xu, Kun Zhou, Jingdong Wang, Dongping Li, and Baining Guo.
\newblock {An Interactive Approach to Semantic Modeling of Indoor Scenes with
  an RGBD Camera}.
\newblock {\em ACM Trans. Graph. (SIGGRAPH ASIA)}, 31(6), 2012.

\bibitem{Sharf2004}
Andrei Sharf, Marc Alexa, and Daniel Cohen-Or.
\newblock {Context-based Surface Completion}.
\newblock {\em ACM Trans. Graph. (SIGGRAPH)}, 23(3), 2004.

\bibitem{Shen2012}
Chao-Hui Shen, Hongbo Fu, Kang Chen, and Shi-Min Hu.
\newblock {Structure Recovery by Part Assembly}.
\newblock {\em ACM Trans. Graph. (SIGGRAPH ASIA)}, 31(6), 2012.

\bibitem{Song2017}
Shuran Song, Fisher Yu, Andy Zeng, Angel~X Chang, Manolis Savva, and Thomas
  Funkhouser.
\newblock Semantic scene completion from a single depth image.
\newblock In {\em Computer Vision and Pattern Recognition (CVPR)}, 2017.

\bibitem{Stutz2018}
David Stutz and Andreas Geiger.
\newblock {Learning 3D shape completion from laser scan data with weak
  supervision}.
\newblock In {\em Computer Vision and Pattern Recognition (CVPR)}. IEEE
  Computer Society, 2018.

\bibitem{Su2017}
Hao Su, Haoqiang Fan, and Leonidas Guibas.
\newblock {A point set generation network for 3D object reconstruction from a
  single image}.
\newblock In {\em Computer Vision and Pattern Recognition (CVPR)}, 2017.

\bibitem{Sung2015}
Minhyuk Sung, Vladimir~G Kim, Roland Angst, and Leonidas Guibas.
\newblock {Data-driven structural priors for shape completion}.
\newblock {\em ACM Trans. Graph. (SIGGRAPH ASIA)}, 34(6), 2015.

\bibitem{Andrea2011}
Andrea Tagliasacchi, Matt Olson, Hao Zhang, Ghassan Hamarneh, and Daniel
  Cohen-Or.
\newblock {VASE: Volume-Aware Surface Evolution for Surface Reconstruction from
  Incomplete Point Clouds}.
\newblock {\em Computer Graphics Forum}, 30(5), 2011.

\bibitem{Tatarchenko2017}
M. Tatarchenko, A. Dosovitskiy, and T. Brox.
\newblock {Octree generating networks: efficient convolutional architectures
  for high-resolution 3D outputs}.
\newblock In {\em International Conference on Computer Vision (ICCV)}, 2017.

\bibitem{Wang2017}
Peng-Shuai Wang, Yang Liu, Yu-Xiao Guo, Chun-Yu Sun, and Xin Tong.
\newblock {O-CNN: Octree-based convolutional neural networks for 3D shape
  analysis}.
\newblock {\em ACM Trans. Graph. (SIGGRAPH)}, 36(4), 2017.

\bibitem{Wang2018a}
Peng-Shuai Wang, Chun-Yu Sun, Yang Liu, and Xin Tong.
\newblock {Adaptive O-CNN: A patch-based deep representation of 3D shapes}.
\newblock {\em ACM Trans. Graph. (SIGGRAPH ASIA)}, 2018.

\bibitem{Wang2019}
Yida Wang, David~Joseph Tan, Nassir Navab, and Federico Tombari.
\newblock {ForkNet}: Multi-branch volumetric semantic completion from a single
  depth image.
\newblock In {\em International Conference on Computer Vision (ICCV)}, 2019.

\bibitem{Wilhelms1992}
Jane Wilhelms and Allen Van~Gelder.
\newblock {Octrees for faster isosurface generation}.
\newblock {\em ACM Trans. Graph.}, 11(3), 1992.

\bibitem{Wu2016}
Jiajun Wu, Chengkai Zhang, Tianfan Xue, William~T. Freeman, and Joshua~B.
  Tenenbaum.
\newblock {Learning a probabilistic latent space of object shapes via 3D
  generative-adversarial modeling}.
\newblock In {\em Neural Information Processing Systems (NeurIPS)}, 2016.

\bibitem{Wu2018b}
Jiajun Wu, Chengkai Zhang, Xiuming Zhang, Zhoutong Zhang, William~T Freeman,
  and Joshua~B Tenenbaum.
\newblock Learning shape priors for single-view {3D} completion and
  reconstruction.
\newblock In {\em European Conference on Computer Vision (ECCV)}, 2018.

\bibitem{Wu2018}
Q. Wu, K. Xu, and J. Wang.
\newblock {Constructing 3D CSG models from 3D raw point clouds}.
\newblock {\em Comput. Graph. Forum}, 37(5), 2018.

\bibitem{Wu2015}
Z. Wu, S. Song, A. Khosla, F. Yu, L. Zhang, X. Tang, and J. Xiao.
\newblock {3D ShapeNets: A deep representation for volumetric shape modeling}.
\newblock In {\em Computer Vision and Pattern Recognition (CVPR)}. IEEE, 2015.

\bibitem{Yang2018}
Bo Yang, Stefano Rosa, Andrew Markham, Niki Trigoni, and Hongkai Wen.
\newblock Dense {3D} object reconstruction from a single depth view.
\newblock {\em IEEE Trans. Pattern Anal. Mach. Intell.}, 2018.

\bibitem{Yin2018}
Kangxue Yin, Hui Huang, Daniel Cohen-Or, and Hao Zhang.
\newblock {P2P-NET: Bidirectional Point Displacement Net for Shape Transform}.
\newblock {\em ACM Trans. Graph. (SIGGRAPH)}, 37(4), 2018.

\bibitem{Zhang2018}
Jiahui Zhang, Hao Zhao, Anbang Yao, Yurong Chen, Li Zhang, and Hongen Liao.
\newblock Efficient semantic scene completion network with spatial group
  convolution.
\newblock In {\em European Conference on Computer Vision (ECCV)}, 2018.

\bibitem{Zhang2019}
Pingping Zhang, Wei Liu, Yinjie Lei, Huchuan Lu, and Xiaoyun Yang.
\newblock Cascaded context pyramid for full-resolution {3D} semantic scene
  completion.
\newblock In {\em International Conference on Computer Vision (ICCV)}, 2019.

\bibitem{Zheng2013}
Bo Zheng, Yibiao Zhao, Joey~C. Yu, Katsushi Ikeuchi, and Song-Chun Zhu.
\newblock {Beyond Point Clouds: Scene Understanding by Reasoning Geometry and
  Physics}.
\newblock In {\em Computer Vision and Pattern Recognition (CVPR)}, 2013.

\bibitem{Zhou2011}
Kun Zhou, Minmin Gong, Xin Huang, and Baining Guo.
\newblock Data-parallel octrees for surface reconstruction.
\newblock {\em IEEE. T. Vis. Comput. Gr.}, 17(5), 2011.

\end{thebibliography}
}

\end{document}